\documentclass[letterpaper]{article} 
\usepackage{aaai24}  
\usepackage{times}  
\usepackage{helvet}  
\usepackage{courier}  
\usepackage[hyphens]{url}  
\usepackage{graphicx} 
\usepackage{natbib}  
\usepackage{caption} 
\usepackage{algorithm}
\usepackage{algorithmic}
\usepackage{times}
\usepackage{epsfig}
\usepackage{graphicx}
\usepackage{amsmath}
\usepackage{amssymb}
\usepackage{amsfonts} 
\usepackage{multirow}
\usepackage{tabularx} 
\usepackage{ragged2e} 
\usepackage{booktabs} 
\usepackage{algorithm}
\usepackage{algorithmic}
\usepackage{booktabs}       
\usepackage{amsfonts}       
\usepackage{nicefrac}       
\usepackage{microtype}      
\usepackage{soul}
\usepackage{amsthm}
\usepackage{amsmath}
\usepackage{amssymb}
\usepackage{booktabs}
\usepackage{multirow}
\usepackage{bbding}
\usepackage{overpic}
\usepackage{subfigure}
\usepackage{makecell}
\usepackage{colortbl,xcolor}

\usepackage{newfloat}
\usepackage{listings}

\def\model{VTUDC}
\def\origin{TOLED and POLED}
\def\dataset{PexelsUDC}

\def\fig{Fig.~}
\def\eqn{Eqn.~}
\def\tab{Tab.~}
\def\attention{STFM}
\def\tt{Temporal Transformer}

\def\sblock{Spatial Transformer Block}

\def\etc{\emph{etc. }}

\DeclareCaptionStyle{ruled}{labelfont=normalfont,labelsep=colon,strut=off} 
\floatstyle{ruled}
\newfloat{listing}{tb}{lst}{}
\floatname{listing}{Listing}
\title{Deep Video Restoration for Under-Display Camera}
\author{
    Xuanxi Chen\textsuperscript{\rm 1}, Tao Wang\textsuperscript{\rm 1}, Ziqian Shao\textsuperscript{\rm 1}, Kaihao Zhang\textsuperscript{\rm 2}, Wenhan Luo\textsuperscript{\rm 3}, Tong Lu\textsuperscript{\rm 1}, \\
    Zikun Liu\textsuperscript{\rm 4}, Tae-Kyun Kim\textsuperscript{\rm 5}, Hongdong Li\textsuperscript{\rm 2}
}
\affiliations{
    \textsuperscript{\rm 1}State Key Lab for Novel Software Technology, Nanjing University, China\\
    \textsuperscript{\rm 2}Australian National University, Australia\\
    \textsuperscript{\rm 3}Shenzhen Campus of Sun Yat-sen University, China\\
    \textsuperscript{\rm 4}Samsung Research China-Beijing (SRC-B), China\\
    \textsuperscript{\rm 5}Imperial College London, UK\\

    \{xxchen.nju, taowangzj, super.khzhang, whluo.china, hongdong.li\}@gmail.com, ziqian.shao@outlook.com,\\
    lutong@nju.edu.cn,
    zikun.liu@samsung.com, tk.kim@imperial.ac.uk, 
}

\usepackage{bibentry}

\begin{document}

\maketitle

\begin{abstract}
Images or videos captured by the Under-Display Camera (UDC) suffer from severe degradation, such as saturation degeneration and color shift. 
While restoration for UDC has been a critical task, existing works of UDC restoration focus only on images. UDC video restoration (UDC-VR) has not been explored in the community. In this work, we first propose a GAN-based generation pipeline to simulate the realistic UDC degradation process. With the pipeline, we build the first large-scale UDC video restoration dataset called {\dataset}, which includes two subsets named {\dataset}-T and {\dataset}-P corresponding to different displays for UDC. Using the proposed dataset, we conduct extensive benchmark studies on existing video restoration methods and observe their limitations on the UDC-VR task. To this end, we propose a novel transformer-based baseline method that adaptively enhances degraded videos. The key components of the method are a spatial branch with local-aware transformers, a temporal branch embedded temporal transformers, and a spatial-temporal fusion module. These components drive the model to fully exploit spatial and temporal information for UDC-VR. Extensive experiments show that our method achieves state-of-the-art performance on {\dataset}. The benchmark and the baseline method are expected to promote the progress of UDC-VR in the community, which will be made public.
\end{abstract}

\section{Introduction}
Recently, under-display camera (UDC) technology has made it a popular trend to enable bezel-free and notch-free viewing experiences on devices such as smartphones, TV, laptops, and tablets. However, the display embedded in the front of the camera directly leads to saturation degeneration, color shift, and contrast reduction in the captured images using UDC~\cite{zhou2021image, feng2021removing}. Restoration under UDC has become a newly-defined task that aims to recover visually pleasing results from degraded inputs. 

To address this problem, recent methods focus on image restoration for UDC. For example, \citet{zhou2021image} first analyze the UDC imaging process and propose two real UDC image datasets (TOLED and POLED) by a monitor-camera image system. \citet{feng2021removing} build a synthetic dataset using the measured point spread function to simulate UDC degraded images, which focuses on diffraction artifacts in HDR images. 
Some methods~\citep{sundar2020deep, zhou2021image, kwon2021controllable, feng2021removing, koh2022bnudc, feng2023generating} resort to different deep learning architectures such as dynamic skip connection network~\cite{feng2021removing}, two-branched network~\cite{koh2022bnudc} to recover clear images in an end-to-end manner. Although these image-based methods work well in processing degraded UDC images, they are sub-optimal to video restoration in UDC. The image-based methods do not consider the inherent temporal information in the video. On the other hand, existing video restoration methods that utilize temporal cues are not specifically tailored for UDC restoration tasks, thereby overlooking the unique characteristics of UDC videos. 

\begin{figure}[t] \centering
    \includegraphics[width=0.45\textwidth]{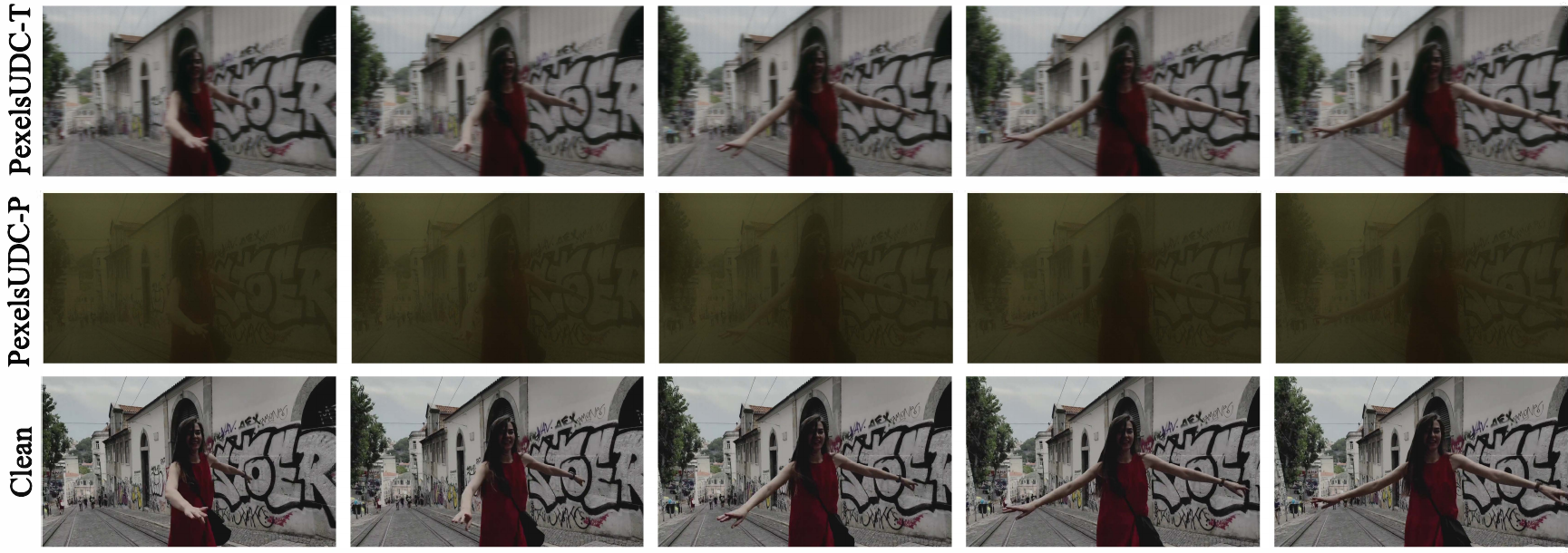}
    \caption{Some examples from our proposed {\dataset}. The bottom row is the clean video sequence. The top and middle rows are sequences sampled from {\dataset}-T and {\dataset}-P, respectively. Zoom in for a better view. }
    \label{fig:udc}
    \vspace{-0.6cm}
\end{figure}

To our best knowledge, few studies have explored UDC Video Restoration (UDC-VR) in the community so far. There are two major unresolved issues that hinder the progress of this task in the community. The first one is the lack of a public and standard benchmark. Till now, there is no benchmark dataset for UDC-VR tasks, which makes it difficult to understand and explore this task in depth. In addition, there exist UDC image datasets, while their scale is small and inherently lacks temporal cues. The second is the lack of baseline algorithms for this task. When a new method is proposed, the authors have to retrain other video restoration methods to achieve a fair comparison, which is effort-demanding.

To advance the development of UDC-VR, we make the first attempt to investigate the UDC-VR problem in this work. To begin with, we propose an automatic pipeline called the UDC video generation pipeline to establish a large-scale video benchmark for UDC-VR. 
Specifically, the proposed pipeline contains four stages: \textit{video collection}, \textit{manual filtering}, \textit{GAN-based UDC video generation}, and \textit{artifact elimination and selection}. 
In our pipeline, the first two phases are used to collect high-quality videos, and the latter two phases are designed to synthesize the corresponding UDC videos. Among them, the GAN-based UDC video generation is the core stage. We propose a generative model to learn the realistic degradation of UDC from the datasets TOLED and POLED~\cite{zhou2021image} which are collected in real scenes.
On the basis of our video generation pipeline, we build the first UDC video dataset called {\dataset} for UDC-VR.  {\dataset} consists of two subsets named {\dataset}-T and {\dataset}-P related to different displays for UDC. Each subset contains $160$ videos, with $140$ videos allocated for training and $20$ videos designated for testing purposes. Each individual video comprises $100$ successive frames, all with a resolution of $1280\times720$. We show some examples in Fig.~\ref{fig:udc}.

Based on our {\dataset}, we first conduct extensive benchmark studies on existing mainstream image/video restoration methods and then we propose a new UDC-VR baseline method (Video Transformer for UDC) called {\model} for the UDC-VR task. {\model} is a transformer-based network with two branches. Specifically, {\model} first adopts a spatial branch and a temporal branch in parallel to extract spatial and temporal cues from the input video, respectively. Then the extracted spatial-temporal information is integrated and enhanced by a spatial-temporal fusion module ({\attention}), which helps the network to dynamically emphasize the complementarity between spatial and temporal information.
After that, the fused feature will pass through stacked spatial transformer blocks to be enhanced again. Finally, the feature is projected to the same size as the reference frame by a convolution with the PixelShuffle~\cite{shi2016real} operation, and the restored reference frame is obtained by residual connection with the reference frame. Experimental results show that {\model} achieves state-of-the-art results on both {\dataset}-T and {\dataset}-P. 

In summary, our main contributions are threefold: (1) We build a new high-quality UDC dataset for video restoration. To the best of our knowledge, it is the first large-scale UDC video dataset in the literature. In addition, we benchmark five state-of-the-art video restoration methods on the two datasets. It helps understand the potential and limitations of these methods on UDC video restoration. (2) We propose a new transformer-based method, {\model}, for UDC video restoration. For the newly-defined UDC video restoration task, this serves as a baseline method. {\model} leverages the characteristics of UDC videos and thus extracts spatio-temporal features more effectively. (3) Comprehensive experiments on two subsets of our dataset demonstrate that {\model} achieves not only state-of-the-art quantitative performance but also consistently superior perceptual quality.
\section{Related Work}
\noindent \textbf{UDC Image Restoration.~}
\citet{zhou2021image} propose two methods to tackle UDC image restoration on {\origin} dataset, which are a conventional method using Wiener Filter \cite{goldstein1998multistage} and a learning-based method with U-shape architecture. 
DISCNet~\cite{feng2021removing} utilizes dynamic convolution and uses the provided PSFs as extra knowledge to guide the model. 
BNUDC~\cite{koh2022bnudc} is a two-branched DNN that considers high-spatial-frequency and low-spatial-frequency degradation for UDC image restoration. 

\noindent \textbf{UDC Datasets and Simulation.~}
\citet{zhou2021image, zhou2020udc} employ a Monitor-Camera Imaging System (MCIS) system to collect paired and aligned realistic data (TOLED and POLED datasets).
\citet{feng2023generating} specifically design AlignFormer to address misalignment problems in the real collected data.
Since the scale of realistic UDC datasets are usually small, and the real paired UDC data is difficult to obtain, \citep{feng2021removing, feng2022mipi} propose a model-based synthesis pipeline based on measured PSF to generate a large-scale synthetic UDC dataset.
Many works also use GANs to simulate UDC images in {\origin} for data augmentation in UDC image restoration. However, the performance of the simulation is not ideal for generating UDC datasets.
The glass simulation in DAGF-GAN~\cite{sundar2020deep} introduces different noise patterns at the image's edge. 
MPGNet~\cite{zhou2022modular} integrates the realistic noise estimation module in GAN for end-to-end training. However, the noise in the original data set is a fixed pattern which is read-shot noise. Meanwhile, the synthetic results of MPGNet differ from {\origin} in detail. We propose a two-stage UDC synthetic pipeline to learn real UDC degradation, decoupling noise estimation from an end-to-end learning framework. To our knowledge, no publicly available UDC video dataset has been available, and one of our main contributions is proposing the first UDC video dataset {\dataset} based on our UDC synthetic pipeline.

\noindent \textbf{Video Restoration with Transformer.~} Transformer is first introduced by \citet{vaswani2017attention} in natural language processing tasks. Then broad ranges of development based on transformers are introduced to high-level vision tasks~\citep{dosovitskiy2020image, carion2020end, zhu2020deformable, wang2021max, touvron2021training, liu2021swin} and low-level vision tasks~\citep{chen2021pre, liang2021swinir, wang2022uformer}. Transformer has also been introduced for video restoration~\citep{kim2018spatio, liang2022vrt, zheng2022progressive, geng2022rstt, lin2022flow, luo2022bsrt}. Inspired by Swin Transformer~\citep{liu2021swin, liang2021swinir}, more video restoration methods have adopted window-based multi-head attention in recent years. \citet{zheng2022progressive} adopt SwinIR~\cite{liang2021swinir} as their backbone model in the second stage. RSTT~\cite{geng2022rstt} exploits the stacked Swin Transformer Block to solve the video super-resolution problem. FGST~\cite{lin2022flow} modifies the original window-based local self-attention to use the guidance of optical flow to expand the receptive field. BSRT~\cite{luo2022bsrt} incorporates Swin Transformer blocks to better use inter-frame information. 
\vspace{-0.3cm}

\section{PexelsUDC Dataset}
\subsection{UDC Degradation Model} 
As TOLED and POLED are real-world collected paired datasets, thus we try to simulate the degradation existing in these datasets. We follow the degradation model proposed by \citet{zhou2021image} and extend it to the video domain. The degradation model of the $t$-th frame in UDC videos can be defined as:
\begin{equation}\label{eq:degradation_model}
y_{t}=(\gamma * x_{t})\otimes k + n, 
\end{equation}
where $y$ is the degraded video frame, $x$ is the clean frame, $k$ is the point spread function (PSF), $\gamma$ is the scale factor that indicates the attenuation of light intensity, $n$ is additive noise, and $\otimes$ represents the convolution operator. It should be noted that $\gamma$, $k$, and $n$ are usually fixed because they are parameters related to the imaging system.

\subsection{Two-Stage Simulation Pipeline}
\begin{figure}[t]
    \centering
    \includegraphics[width=0.45\textwidth]{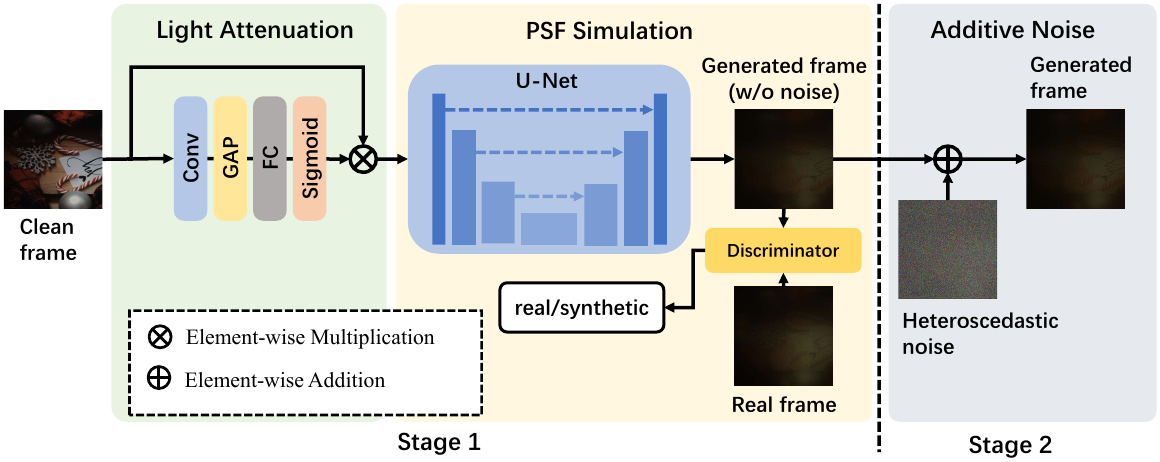}
    \vspace{-0.2cm}
    \caption{The structure of our two-stage simulation pipeline. The first stage is a GAN that learns the light attenuation and PSF. The second stage is adding heteroscedastic noise to the generated frame of the first stage.}
    \label{fig: gan}
    \vspace{-0.5cm}
\end{figure}

Based on the above degradation model, we propose a two-stage generation method to learn the degradation of UDC. The overall architecture of our method is shown in {\fig}\ref{fig: gan}. The details of two-stage of our methods are shown as follows:

\noindent\textbf{Degradation Learning via GAN Model.~}
We adopt GAN to learn the degradation from UDC images in {\dataset}.
For the generator, the main body is a customized U-shaped network \cite{ronneberger2015u}. Inspired by~\cite{zhou2022modular}, we also design a light attenuation module and put it in front of the U-shape network to facilitate the generation. See Supp. Material for more details about the model. For the discriminator, we use PatchGAN discriminator~\cite{isola2017image}, which has been proven to be effective in image-to-image translation fields.

\noindent \textbf{Additive Noise.~}
\citet{zhou2021image} adopt a simple noise model following the commonly used read-shot noise, which can be formulated as a heteroscedastic Gaussian:
\begin{equation}\label{eq:noise}
n(x)\sim N(\mu=0, \sigma^2=\lambda_{read}+\lambda_{shot}x),
\end{equation}
where the mean value is $0$, and the variance is signal-dependent. $\lambda_{read}$ and $\lambda_{shot}$ are parameters related to the camera. 
Due to the fixed noise pattern, we do not include a noise learning module in our GAN. Thus, at the end of stage 1 in {\fig}\ref{fig: gan}, the image generated by GAN does not contain noise. Since $\lambda_{read}$ and $\lambda_{shot}$ are not available, and we do not have pairs of images with and without noise for training, we manually set some candidate values for $\lambda_{read}$ and $\lambda_{shot}$. To determine $\lambda_{read}$ and $\lambda_{shot}$, twenty volunteers were recruited to rate how close the generated images were to real images under each set of parameters. Furthermore, temporal coherence is crucial for synthetic video datasets.
UDC degradation differs from other degradation, such as rain, haze, and snow. 
UDC degradation originates from the imaging system rather than the environment captured in the image, so once we have effectively learned the characteristics of UDC degradation, we possess a "camera lens" for UDC.
Hence, stable GAN outputs can ensure temporal coherence at moderate frame rates. 
We choose moderate videos with moderate frame rates to alleviate this problem.


\subsection{Our {\dataset}}
Based on the UDC simulation pipeline, we propose a high-quality and large-scale benchmark dataset {\dataset} considering two kinds of degradation (TOLED and POLED) for UDC-VR. 
To obtain the source clean videos, we resort to Pexels\footnote{\url{https://www.pexels.com/}}, where all the videos are free to use. 
The ground-truth videos we collected are all real shots taken by different smart devices, which include a variety of content like people, animals, natural scenery, \etc
The FPS of videos in our dataset is relatively moderate, between 25 and 50.
In {\dataset}, there are $140$ pairs of videos for training and $20$ pairs for testing. All the videos are of $1280\times 720$ resolution, and each sequence has $100$ frames. 
See Supp. Material for more information about our {\dataset}.

\begin{figure*}[t]
    \centering
    \includegraphics[width=0.9\textwidth]{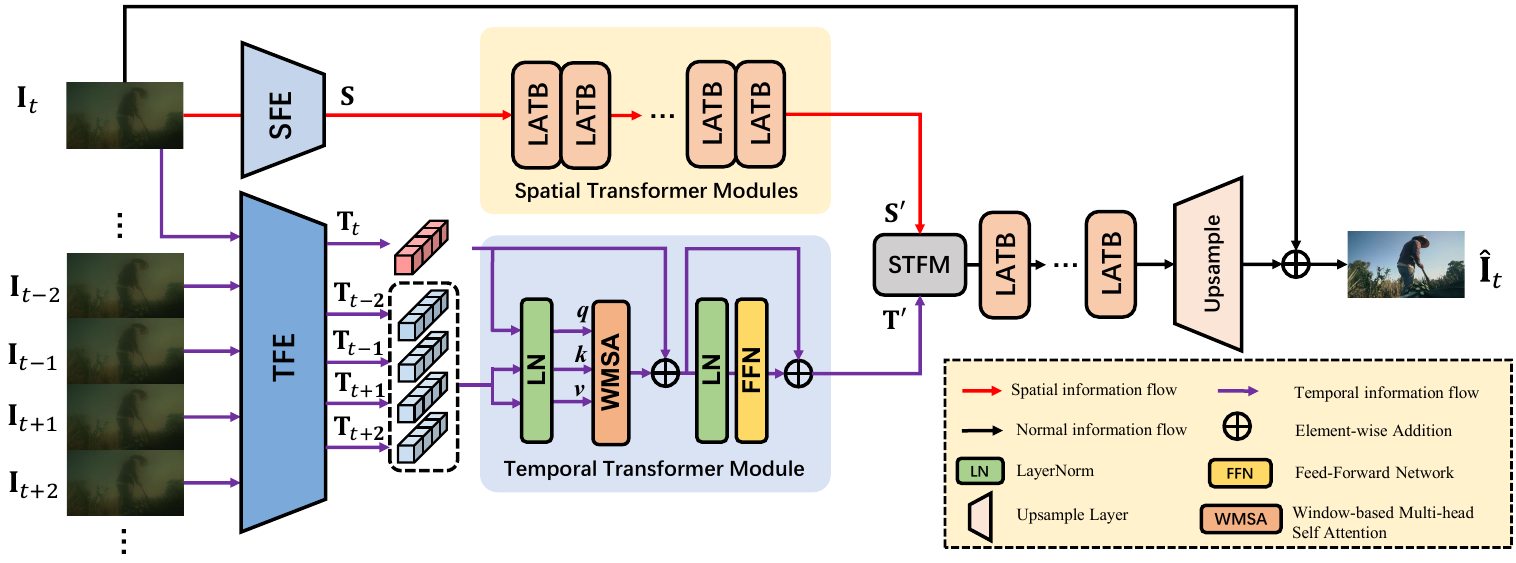}
    \vspace{-0.2cm}
    \caption{The overall structure of proposed {\model}. The reference frame is sent into the spatial branch. It passes through Spatial Feature Extractor (SFE) to get the shallow spatial feature $\mathbf{S}$. Then stacked Local-Aware Transformer Blocks (LATB) are adopted to obtain the spatial feature $\mathbf{S}'$.
    The video sequence is sent into the temporal branch where the feature of each frame is first extracted by Temporal Feature Extractor (TFE). Then the features of frames are fed into Temporal Transformer Module to obtain the temporal feature $\mathbf{T}'$. Then $\mathbf{S}'$ and $\mathbf{T}'$ are fused together by the Spatial-Temporal Fusion Module (STFM) and enhanced by following stacked LATBs. The structure of STFM and LATBs are shown in {\fig}\ref{fig: elements}.}
    \label{fig: model}
    \vspace{-0.5cm}
\end{figure*}

\section{Proposed Method}
As shown in {\fig}\ref{fig: model}, {\model} is a two-branch transformer-based method. Given $2N+1$ degraded frames, the parallel \textbf{spatial and temporal branches} first process the input reference frame and video frames to capture spatial and temporal information. Then, the features produced by these two branches are fused and enhanced via a \textbf{spatial-temporal fusion module}. Finally, the features are recovered into the clear reference frame $N$. We discuss these core components of {\model} as follows.

\subsection{Spatial Branch}\label{sec: spatial_branch}
Through the comprehensive analysis of UDC image degradation, we find that blur and noise are the main degradation factors of the UDC restoration task. Considering the local information is helpful for blur and noise removal~\cite{zhou2022modular}, we thus resort to using the local-aware transformer block as one basic unit to build the spatial branch of the network. As illustrated in Fig.~\ref{fig: model}, the spatial branch contains one shallow feature extractor and several stacked local-aware transformer blocks. The extractor is used to extract shallow features from the input reference frame, and the stacked local-aware transformer blocks are designed to guide the network to focus on local information for UDC restoration. As the local-aware transformer block is our core component, we discuss it as follows.

\begin{figure}[t]
    \centering
    \includegraphics[width=0.45\textwidth]{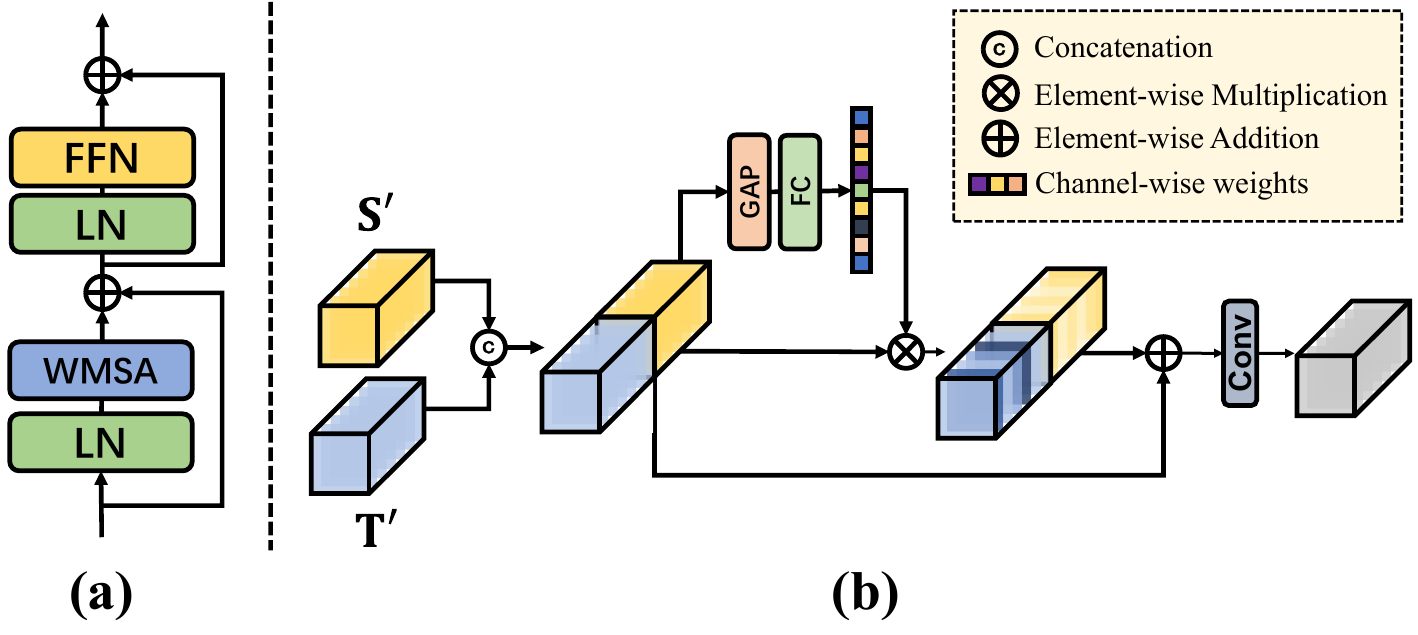}
    \vspace{-0.2cm}
    \caption{(a) Local-Aware Transformer Block (LATB). (b) Spatial-Temporal Fusion Module (STFM). LN is the LayerNorm. FFN is the feed-forward network. WMSA is window-based multi-head self-attention.}\label{fig: elements}
    \vspace{-0.5cm}
\end{figure}

Recent transformer networks have shown significant performance improvements on low-level vision tasks. However, the transformer structure usually conducts global self-attention, which may not be suitable for the UDC video restoration task. Thus, motivated by~\citet{liu2021swin}, we introduce the window-based mechanism in the self-attention to build a local-aware transformer block. Compared with the original ViT~\cite{dosovitskiy2020image}, window-based self-attention focuses more on local information and reduces the computational complexity, which can better help blur and noise removal for the UDC restoration task. Suppose the input feature is $\mathbf{X}\in \mathbb{R}^{C\times H\times W}$. In the local-aware transformer block, $\mathbf{X}$ is first divided into non-overlapping windows: $\mathbf{X} \rightarrow \{\mathbf{X}_1, \mathbf{X}_2, ..., \mathbf{X}_n\}, n=HW/M^2$ with the window size of $M\times M$, where $n$ is the total number of windows. In addition, we apply the learnable absolute position embedding on feature $\mathbf{X}_{i}$ of each window. Then the feature maps of each window are flattened and transposed into the shape of $\mathbb{R}^{M^2\times C}$. After that, we perform multi-head self-attention in each window independently. For each window, let the head number be $k$, and the head dimension be $d=C/k$, the query, key, and value under the $k$-th head self-attention are generated as follows:
\begin{equation}
\mathbf{Q}_{i}=\mathbf{X}_{i}\mathbf{W}_{Q},~~ \mathbf{K}_{i}=\mathbf{X}_{i}\mathbf{W}_{K},~~ \mathbf{V}_{i}=\mathbf{X}_{i}\mathbf{W}_{V}, 
\end{equation}
where $\mathbf{W}_{Q}$, $\mathbf{W}_{K}$, $\mathbf{W}_{V} \in \mathbb{R}^{C\times d}$ represent the projection matrices of query, key, and value, respectively. The formulation for attention is:
\begin{equation}\label{eq:attention}
    \text{Attention}(\mathbf{Q}, \mathbf{K}, \mathbf{V}) = \text{Softmax}(\frac{\mathbf{Q}\mathbf{K^{\emph{T}}}}{\sqrt{d}})V.
\end{equation}
Next, the feature passes through the feed-forward network (FFN). FFN consists of two fully-connected (FC) layers, each layer is followed by a GELU activation function. The whole process of the local-aware block is formulated as follows:
\begin{equation}
\begin{split}
    &\mathbf{X}'=\text{MHSA}(\text{FFN}(\text{LN}(\text{PE}(\mathbf{X})))) + \mathbf{X},\\
    &\mathbf{X}^{\prime\prime}=\text{FFN}(\text{LN}(\mathbf{X}^{\prime})) + \mathbf{X}^{\prime},
\end{split}
\end{equation}
where $\text{LN}$ indicates the LayerNorm~\cite{ba2016layer} layer, and $\text{PE}$ is the patch embedding. We illustrate the structure of the local-aware block in {\fig}\ref{fig: elements} (a). In addition, we also utilize the shifted window mechanism in these stacked local-aware transformer blocks to further enhance the long-range dependency modeling capacities of the spatial branch.

\begin{table*}[t]
    \centering
    \newcommand{\Frst}[1]{\textcolor{red}{\textbf{#1}}}
    \newcommand{\Scnd}[1]{\textcolor{blue}{\textbf{#1}}}
    \newcommand{\psnr}{PSNR$\uparrow$}
    \newcommand{\ssim}{SSIM$\uparrow$}
    \newcommand{\lpips}{LPIPS$\downarrow$}
    \newcommand{\dists}{DISTS$\downarrow$}
    \caption{The effectiveness of our GAN-based UDC generation method. We train DWFormer~\cite{zhou2022modular} on three different datasets: Original data, Data-MPGNet, and Data-Ours. The original data refers to the testing sets of TOLED and POLED, while Data-MPGNet and Data-Ours are generated by MPGNet and our method, respectively. The best and second-best results are marked in bold and underlined respectively.} 
    \vspace{-0.2cm}
    \begin{tabular}{|l|*{2}{c}*{2}{c}|*{2}{c}*{2}{c}|}
        \cline{1-9}
        \multirow{2}*{Training Dataset} & \multicolumn{4}{c|}{\textbf{TOLED-Val}} & \multicolumn{4}{c|}{\textbf{POLED-Val}} \\
        \cline{2-9}
        & \psnr & \ssim & \lpips & \dists & \psnr & \ssim & \lpips & \dists \\
        \cline{1-9}
        Original data & \textbf{31.52} & \textbf{0.8723} & \textbf{0.3605} & \textbf{0.1920} & \textbf{21.23} & \textbf{0.6880} & \textbf{0.5391} & \textbf{0.3417} \\
        Data-MPGNet & 28.97 & 0.8269 & 0.3830 & 0.2912 & 16.86 & 0.5780 & 0.5580 & 0.3721 \\
        Data-Ours & \underline{30.49} & \underline{0.8568} & \underline{0.3780} & \underline{0.2203} & \underline{20.77} & \underline{0.6340} & \underline{0.5498} & \underline{0.3702} \\
        \cline{1-9}
    \end{tabular}
    \vspace{-0.2cm}
    \label{tab: validation}
\end{table*}
\begin{figure*}[htbp]
    \centering
    \includegraphics[width=0.90\textwidth]{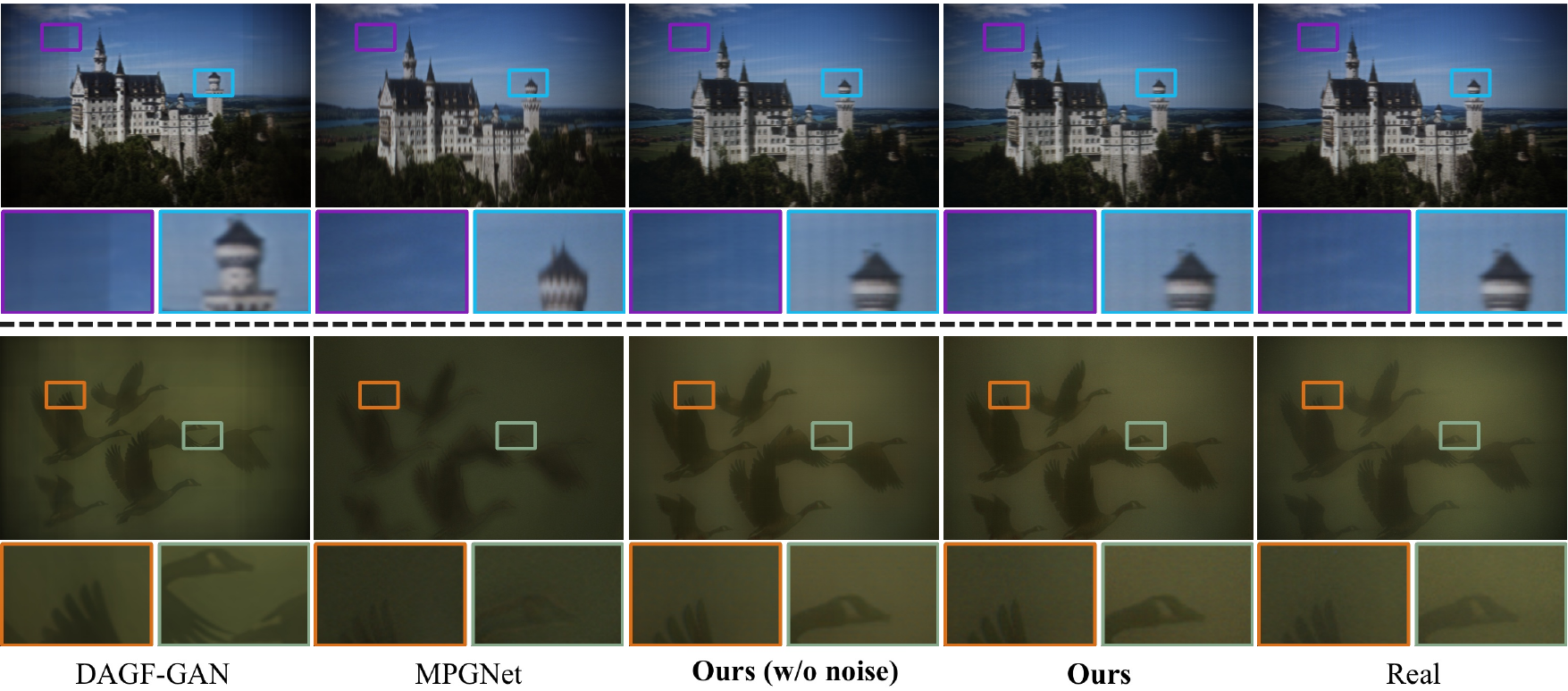}
    \vspace{-0.2cm}
    \caption{Generation comparison on TOLED and POLED validation sets. The top row shows TOLED validation results, and the bottom row shows POLED validation results. DAGF-GAN~\cite{sundar2020deep} is trained on original images from DIV2K~\cite{agustsson2017ntire} which are misaligned with images in {\origin}. Official pre-trained models of DAGF-GAN and MPGNet are adopted.}
    \label{fig: gen}
    \vspace{-0.5cm}
\end{figure*}

\subsection{Temporal Branch}\label{sec: temporal_transformer}
In contrast to UDC image restoration, UDC-VR is more challenging because it is not clear how to utilize the inherent temporal information in successive video frames to assist the restoration process. Thus, we propose a temporal branch in the network to exploit the temporal information from input videos. Specifically, the temporal branch consists of a U-shaped feature extractor composed of a local-aware transformer block and a temporal transformer module. The feature extractor is used to transform input frames into features, and the temporal transformer module is adopted to seek complementary sharp information existing in neighboring frames. The structure of {\tt} is shown in {\fig}\ref{fig: model}. Specifically, given the feature maps $\mathbf{T} \in \mathbb{R}^{K\times C\times H'\times W'}$ after feature extraction, the feature maps are split into non-overlapping windows with the size of $M_{t}\times M_{t}$. Then the feature maps of each window $j$ in frame $i$ will be flattened and transposed to obtain $\mathbf{T}_{i, j}\in \mathbb{R}^{M_{t}^2\times C}$. In this way, features of every frame $\mathbf{T}_{i}$ is in the shape of $\mathbb{R}^{M_{t}^2\times \frac{H'W'}{M_t^2}\times C}$. The feature of the reference frame $\mathbf{T}_{t}$ is sent into the multi-head self-attention module as the query, and the features of neighboring frames are treated as the key and value. Then we perform the multi-head self-attention described in {\eqn}\ref{eq:attention} to derive the final temporal feature $\mathbf{T'}\in \mathbb{R}^{C\times H'\times W'}$. Compared to previous works using optical flow \cite{lin2022flow} or deformable convolution \cite{wang2019edvr} to model temporal information, our temporal transformer is simple and can directly improve restoration performance.

\subsection{Spatial-Temporal Fusion Module}\label{sec: stfm}
After obtaining information from the spatial and temporal domains, we need to fuse them together. It is necessary to choose between spatial and temporal information and emphasize the useful information in the features of both spatial and temporal parts. Therefore, we propose a spatial-temporal fusion module, where channel attention is adopted to re-weight the features from two branches. The structure of {\attention} is shown in {\fig}\ref{fig: elements} (b).

The feature map $\mathbf{S'}\in \mathbb{R}^{C\times H'\times W'}$ from the spatial branch and $\mathbf{T'}\in \mathbb{R}^{C\times H'\times W'}$ from the temporal branch are first concatenated together which is denoted by $\mathbf{F}\in \mathbb{R}^{2C\times H'\times W'}$. Then $\mathbf{F}$ passes through the global average pooling (GAP) and a fully-connected (FC) layer to derive the weights $\mathbf{W}\in \mathbb{R}^{2C\times 1\times1}$ for every channel in $\mathbf{F}$ which is a channel attention (CA) operation. Then we multiply $\mathbf{F}$ and $\mathbf{W}$ and make the residual connection with $\mathbf{F}$. Then the features are downsampled to obtain the fused feature $\mathbf{F}^{\prime}\in \mathbb{R}^{C\times H'\times W'}$. This process can be formulated as: 
\begin{equation}
\begin{split}
    & \text{CA}(\mathbf{F}) = \text{FC}(\text{GAP}(\mathbf{F}))*\mathbf{F},\\
    & \mathbf{F}' = \text{Conv}(\text{CA}(\mathbf{F}))+\mathbf{F}.
\end{split}
\end{equation}
Then $\mathbf{F}'$ passes through the stacked local-aware transformer blocks to be enhanced. Finally, $\mathbf{F}$ is projected to the same size as the reference frame, taken as residual, and added to itself to derive the restored reference frame.

\begin{table*}[t]
    \centering
    \newcommand{\Frst}[1]{\textcolor{red}{\textbf{#1}}}
    \newcommand{\Scnd}[1]{\textcolor{blue}{\textbf{#1}}}
    \newcommand{\psnr}{PSNR$\uparrow$}
    \newcommand{\ssim}{SSIM$\uparrow$}
    \newcommand{\lpips}{LPIPS$\downarrow$}
    \newcommand{\dists}{DISTS$\downarrow$}

    \caption{UDC-VR results on {\dataset}. The best and second-best results are marked in bold and underlined, respectively.}
    \vspace{-0.2cm}
    \scalebox{0.95}{
        \begin{tabular}{|l|*{2}{c}*{2}{c}|*{2}{c}*{2}{c}|}
            \cline{1-9}
            \multirow{3}*{Methods}&\multicolumn{8}{c|}{{\dataset}}\\
            \cline{2-9}
            & \multicolumn{4}{c|}{{\dataset}-T} & \multicolumn{4}{c|}{{\dataset}-P} \\
            \cline{2-9}
            & \psnr & \ssim & \lpips & \dists & \psnr & \ssim & \lpips & \dists \\
            \cline{1-9}
            DWFormer & 31.61 & 0.8903 & 0.2610 & 0.1185 & 24.91 & 0.7901 & 0.3879 & 0.1729 \\
            DAGF   & 30.75 & 0.8693 & 0.2636 & 0.1226 & 27.99 & 0.8321 & 0.3580 & 0.1667 \\
            \cline{1-9}
            DBN    & 31.70 & 0.8942 & 0.2474 & 0.1087 & 27.96 & \underline{0.8552} & \underline{0.3178} & \underline{0.1349} \\
            EDVR   & \underline{32.33} & 0.8778 & 0.2678 & \underline{0.0936} & 27.70 & 0.8122 & 0.3846 & 0.1486\\
            DBLRNet  & 32.21 & 0.8999 & 0.2462 & 0.1016 & \textbf{28.03} & 0.8441 & 0.3533 & 0.1424 \\
            IFI-RNN & 32.29 & \underline{0.9022} & \underline{0.2362} & 0.1067 & 27.68 & 0.8530 & 0.3278 & 0.1443 \\
            FGST  & 31.75 & 0.8939 & 0.2592 & 0.1164 & 24.92 & 0.7790 & 0.4358 & 0.1743 \\
            \cline{1-9}
            \textbf{Ours} & \textbf{33.61} & \textbf{0.9187} & \textbf{0.2016} & \textbf{0.0856} & \underline{28.01} & \textbf{0.8645} & \textbf{0.2966} & \textbf{0.1255}\\
            \cline{1-9}
        \end{tabular}
    }
    \vspace{-0.2cm}
    \label{tab: main_reslut}
\end{table*}

\begin{figure*}[t] \centering
    \includegraphics[width=0.95\textwidth]{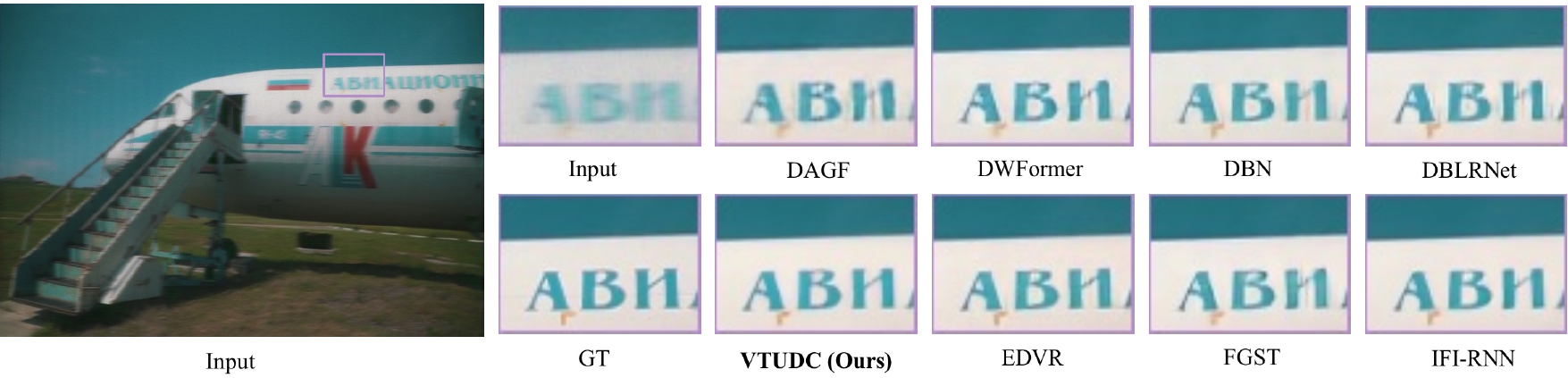}
    \vspace{-0.2cm}
    \caption{Exemplar results on the subset {\dataset}-T from {\dataset}. The image restored by {\model} is clearer and contains fewer artifacts. Zoom in for a better view.}
    \label{fig: t_results}
\end{figure*}

\begin{figure*}[t] \centering
    \includegraphics[width=0.95\textwidth]{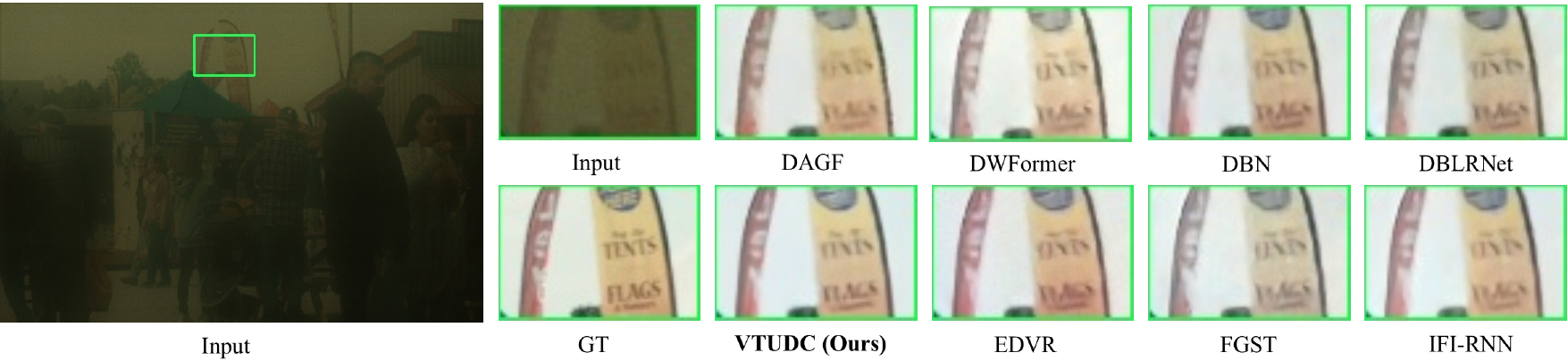}
    \vspace{-0.2cm}
    \caption{Exemplar results on the subset {\dataset}-P from {\dataset}. The details restored by {\model} are sharper and the noise is removed. Zoom in for a better view.}
    \label{fig: p_results}
    \vspace{-0.3cm}
\end{figure*}

\section{Experiments}
\subsection{Experimental Settings}\label{sec: settings}
We train our GAN-based model on both TOLED and POLED \cite{zhou2021image} separately. As to the data augmentation, random cropping ($512\times 512$ patches) and flipping (both horizontal and vertical)~are used. 
The generator and discriminator are optimized by Adam \cite{kingma2014adam} where $\beta_1=0.5$ and $\beta_2=0.999$. The initial learning rate is set to $4\times 10^{-4}$ and the batch size is set to $1$. 

\noindent \textbf{UDC-VR.~}{\model} receives 5 consecutive frames as input. For data augmentation, random cropping ($256\times 256$ patches) and flipping (both horizontal and vertical) are adopted. All the modules are trained with the Adam optimizer where $\beta_1=0.9$ and $\beta_2=0.99$. The initial learning rate is set to $4\times 10^{-4}$ and the batch size is set to $2$. We train {\model} using both Charbonnier loss~\cite{charbonnier1994two} and Perceptual loss~\cite{johnson2016perceptual} with the weight 1 and $1\times10^{-4}$, respectively. {\model} are trained with four NVIDIA V100 GPUs with 32G memory.

\noindent \textbf{Evaluation metrics.~}
For both UDC image generation and UDC-VR, we adopted PSNR, SSIM~\cite{wang2004image}, LPIPS~\cite{zhang2018unreasonable} and DISTS~\cite{ding2020image} to make a comprehensive comparison. PSNR and SSIM are pixel-wise metrics. LPIPS and DISTS are perceptual metrics. 

\subsection{UDC Generation Analysis}
We choose two existing UDC simulation methods for comparison, which are DAGF-GAN \cite{sundar2020deep} and MPGNet \cite{zhou2022modular}.
We test all UDC generation models on TOLED and POLED validation sets.
Quantitative results (See Supp. Material) show that our method outperforms the existing UDC simulation methods on all metrics.
The visualization results are demonstrated in {\fig}\ref{fig: gen}. For both TOLED and POLED, the images generated by DAGF-GAN show different patterns at the edges compared to real degraded images, and there is no evident noise in the images as DAGF-GAN does not consider noise.
MPGNet has a module specifically for noise and thus achieves better performance than DAGF-GAN. 
However, MPGNet still suffers from the following problems.
First, it fails to generate the periodic bands in TOLED. 
Second, the degradation learned by MPGNet deviates from the true TOLED in detail. 
As shown in {\fig}\ref{fig: gen}, the spire of the tower generated by MPGNet shows different patterns from the real one evidently. Our GAN-based model outperforms the other two methods, where the generated images are much closer to the real UDC image. In particular, we can delineate fine details such as diffraction haze in POLED and banded periodic noise in TOLED. 

In addition, we conduct training of DWFormer~\cite{zhou2022modular} using three distinct settings. Firstly, we use the original testing set of TOLED and POLED as the original data and generate UDC images by using GT images from the original data through both MPGNet and our method. Following this, DWFormer is trained separately using these three sets of data. Then, we evaluate the trained DWFormers on the validation set of TOLED and POLED. 
Tab.~\ref{tab: validation} demonstrates that the identical model trained on the data generated by our method outperforms the model trained on the synthesized data utilizing the MPGNet method. Moreover, it almost approaches the results of training with genuine UDC images, showcasing the effectiveness of our approach.


\subsection{Results for UDC-VR}
Since there is no previous method for UDC video restoration, we select two UDC image restoration methods and several representative video restoration methods for comparison. For UDC image restoration methods, we select \textbf{DAGF}~\cite{sundar2020deep} and \textbf{DWFormer}~\cite{zhou2022modular}. For video methods , we choose \textbf{DBN}~\cite{su2017deep}, \textbf{DBLRNet}~\cite{zhang2018adversarial} \textbf{EDVR}~\cite{wang2019edvr}, \textbf{IFI-RNN}~\cite{nah2019recurrent} and \textbf{FGST}~\cite{lin2022flow}. 
For a fair comparison, all methods are retrained using the official settings, with the only variation being the number of training epochs, which is set to 500 for all methods.

The quantitative results are reported in {\tab}\ref{tab: main_reslut}. 
On {\dataset}-T, {\model} achieves the best performance of restoration under all metrics. Specifically, {\model} has an advance of 1.28dB and 0.0165 than the second-best methods in terms of PSNR and SSIM. On {\dataset}-P, {\model} achieves the highest SSIM value and the second-best PSNR, and it ranks first on both LPIPS and DISTS. The distance between {\model} and the best DBLRNet~\cite{zhang2018adversarial} in PSNR is only 0.02dB. The visualization results are shown in {\fig}\ref{fig: t_results} and {\fig}\ref{fig: p_results}. For {\dataset}-T, {\model} achieves better performance where the letters are well restored. In {\dataset}-P, the results of {\model} are clearer. In addition, {\model} restores faithful details and removes the noise compared to other methods. In addition, to verify the temporal stability of the videos restored by our method, we used a pre-trained SPyNet \cite{ranjan2017optical} to calculate the optical flow of the model's predicted frames and the ground truth frames. Then we calculate the endpoint error (EPE) between the optical flow of the restored video and the ground truth video. We chose two representative methods (video-based method DBLRNet and image-based method DWFormer) for comparison. The experiment results in Tab.~\ref{tab: epe} show that our method can achieve better temporal stability. 

\begin{table}[t]
    \centering
    \caption{The EPE results. The lower indicator is better.}
    \vspace{-0.3cm}
    \scalebox{0.80}{
    \setlength{\tabcolsep}{4mm}
    \begin{tabular}{|c|ccc|}
        \cline{1-4}
        Method & Ours & DBLRNet & DWFormer  \\
        \cline{1-4}
        TOLED    & \textbf{1.0578} & 1.2375 & 1.2820 \\ 
        POLED    & \textbf{1.9387} & 2.1098 & 3.5435 \\
        \cline{1-4}
        \end{tabular}
    }
    \label{tab: epe}
    \vspace{-0.3cm}
\end{table}
\begin{table}[tb]\centering
    \caption{Ablation studies on branches of VTUDC.}
    \vspace{-0.3cm}
    \scalebox{0.80}{
    \setlength{\tabcolsep}{2.9mm}{
    \begin{tabular}{|cc|c|}
    \cline{1-3}
       Spatial branch & Temporal Branch & PSNR$\uparrow$/SSIM$\uparrow$\\
        \cline{1-3}
    $\checkmark$ &              &31.96/0.8903\\
                 & $\checkmark$ &32.15/0.9045\\
    $\checkmark$ & $\checkmark$ &\textbf{33.59}/\textbf{0.9193}\\
    \cline{1-3}
    \end{tabular}}
    }
    \label{tab: abl_branch}
      \vspace{-0.6cm}
\end{table}


\subsection{Ablation Studies} 
For ablation experiments, we train all models for $300$ epochs on {\dataset}-T, and other settings are not changed.

\textbf{Two branches vs. Single branch.} We explore the effect of the two-branch structure. We remove only one branch at a time and add numbers of {\sblock}s to keep the parameters similar to the two-branch one. The results are reported in {\tab}\ref{tab: abl_branch}. It is observed that using both spatial and temporal information is important for UDC-VR. 

\textbf{Spatial-temporal fusion module}.
In this part, we explore the way in which the information of the two branches are fused and the results are shown in {\tab}\ref{tab: abl_fusion}. We use two other modules to replace STFM, which are directly adding the information from the two branches (denoted as Addition) and concatenating first and then passing through the convolutional layer (denoted as Concat). The results show that {\attention} can improve the performance of {\model} for 0.56 dB compared with using Concat operation.
\begin{table}[th]\centering
     \caption{Ablation studies on feature fusion.}
    \vspace{-0.3cm}
    \scalebox{0.8}{
    \setlength{\tabcolsep}{6mm}{
    \begin{tabular}{|c|ccc|}
    \cline{1-4}
       Fusion way & Addition & Concat & STFM\\
    \cline{1-4}
    PSNR$\uparrow$ & 32.45 & 33.03   & \textbf{33.59}\\ 
    SSIM$\uparrow$ & 0.9036 & 0.9104  & \textbf{0.9193}\\
    \cline{1-4}
    \end{tabular}}
    }
    \label{tab: abl_fusion}
\vspace{-0.3cm}
\end{table}
\begin{table}[tb]
    \centering
    \caption{Ablation studies on the selection modes of query, key, and value in Temporal Transformer.}
    \vspace{-0.4cm}
    \scalebox{0.8}{
    \setlength{\tabcolsep}{6mm}{
    \begin{tabular}{|c|ccc|}
    \cline{1-4}
    Mode & Model (1) & Model (2) & Ours\\
    \cline{1-4}
    PSNR$\uparrow$ & 33.12  & 32.99  & \textbf{33.59}  \\
    SSIM$\uparrow$ & 0.9126 & 0.9128 & \textbf{0.9120} \\
    \cline{1-4}
    \end{tabular}}
    }
    \label{tab: temp}
    \vspace{-0.4cm}
\end{table}

\textbf{The selection mode of query, key, and value}.
We investigate how query, key, and value are selected in Temporal Transformer. We used two other selection modes for comparison: 1) $\mathbf{F}_t$ as query, $\mathbf{F}_{t+i}$ as key and value. 2) $\mathbf{F}_{t+i}$ as query, key and value. The ablated results are shown in {\tab}\ref{tab: temp}, indicating that the validity of information extracted by both settings is weaker than the way we adopt currently, and the latter is inferior to the former.
This suggests that performance will reduce when the feature of the reference frame is involved in key and value, verifying the effectiveness of our Temporal Transformer.


\textbf{Window size for Temporal Transformer}.
We change the window size in {\tt}. {\model} works best with a $4\times 4$ window ($1\times1$: 32.90/0.9108, $2\times2$: 32.98/0.9130, $4\times4$: \textbf{33.59/0.9193}, $8\times8$: 33.05/0.9135 in PSNR/SSIM). A too-small window may ignore some useful information embedded in neighboring frames, while a too-large window may involve interference information from neighboring frames when motion changes greatly.

\section{Conclusion}
In this paper, we explore the UDC video restoration task which has not been studied in the community. We propose a UDC video generation pipeline where high-quality UDC videos can be generated. Based on this pipeline, we construct the first dataset for UDC video restoration. For this new task, we design a transformer-based method ({\model}) of a two-branch structure to tackle it. 
With the exploitation of both spatial and temporal cues, {\model} achieves better performance than other video restoration methods on the proposed dataset both quantitatively and qualitatively. 
All datasets and models will be released to advance the development of UDC-VR within the community.

\bibliography{aaai24}

\end{document}